\documentclass[letterpaper, 10 pt, journal, twoside]{IEEEtran}

\usepackage{url}
\usepackage{cite}
\usepackage{amsmath,amssymb,amsfonts}
\usepackage{algorithmic}
\usepackage{graphicx}
\usepackage{textcomp}
\usepackage[ruled, linesnumbered, vlined, algo2e]{algorithm2e}
\usepackage{caption}
\usepackage{float} 
\usepackage{graphbox}
\usepackage{tcolorbox}
\usepackage[export]{adjustbox}
\usepackage{subcaption}
\usepackage{bibentry}
\usepackage{multirow} 
\usepackage{makecell}
\usepackage{threeparttable}

\usepackage{algorithm}
\usepackage{bm}
\usepackage{booktabs}
\usepackage[hidelinks]{hyperref}

\usepackage{enumitem}

\usepackage[table]{xcolor}
\definecolor{purple2}{HTML}{EBEAFD}
\definecolor{purple1}{HTML}{DDC9FD}
\usepackage{arydshln}

\allowdisplaybreaks[4]

\begin{document}

\title{DPNet: Doppler LiDAR Motion Planning\\for Highly-Dynamic Environments}

\author{
Wei Zuo$^{1}$, Zeyi Ren$^{1}$, Chengyang Li$^{1}$, Yikun Wang$^{1}$, Mingle Zhao$^{3}$, Shuai Wang$^{2,\dagger}$,\\Wei Sui$^{4}$, Fei Gao$^{5}$, Yik-Chung Wu$^{1,\dagger}$, and Chengzhong Xu$^{3}$
\vspace{-0.3in}
\thanks{Project Page: \href{https://github.com/UUwei-zuo/DPNet}{\tt\footnotesize https://github.com/UUwei-zuo/DPNet}}.
\thanks{Manuscript received: November 27, 2025 ; Revised: February 28, 2026; Accepted: April 9, 2026. This paper was recommended for publication by Editor Ashis Banerjee upon evaluation of the Associate Editor and Reviewers’ comments.
This work was supported by National Natural Science Foundation of China (Grant No. 62371444), TUBITAK-CAS Bilateral Cooperation Program (Grant No. 321GJHZ2025118MI), MOST and The Science and Technology Development Fund (FDCT) of Macau SAR (File no. 0074/2025/AMJ), Shenzhen Science and Technology Program (Grant No. RCYX20231211090206005, JCYJ20241202124934046), and SIAT Original Exploration Program.} 
\thanks{$^{1}$ Wei Zuo, Zeyi Ren, Chengyang Li, Yikun Wang, and Yik-Chung Wu are with The University of Hong Kong, Hong Kong 
({\tt\footnotesize \{\href{mailto:zuowei@eee.hku.hk}{\color{black}zuowei}, \href{mailto:renzeyi@eee.hku.hk}{\color{black}renzeyi},\\\href{mailto:ykwang@eee.hku.hk}{\color{black}ykwang},\href{mailto:ycwu@eee.hku.hk}{\color{black}ycwu}\}@eee.hku.hk} and {\tt\footnotesize \href{mailto:chengyangli@connect.hku.hk}{\color{black}chengyangli}@connect.hku.hk}).}%
\thanks{$^{2}$ Shuai Wang is with Shenzhen Institutes of Advanced Technology, Chinese Academy of Sciences, China ({\tt\footnotesize \href{mailto:s.wang@siat.ac.cn}{\color{black}s.wang}@siat.ac.cn}).}
\thanks{$^{3}$ Mingle Zhao and Chengzhong Xu are with University of Macau, Macau ({\tt\footnotesize \href{mailto:zhao.mingle@connect.um.edu.mo}{\color{black}zhao.mingle}@connect.um.edu.mo} and {\tt\footnotesize \href{mailto:czxu@um.edu.mo}{\color{black}czxu}@um.edu.mo}).}
\thanks{$^{4}$ Wei Sui is with D-Robotics, China ({\tt\footnotesize \href{mailto:wei.sui@d-robotics.cc}{\color{black}wei.sui}@d-robotics.cc}).} 
\thanks{$^{5}$ Fei Gao is with Zhejiang University, China ({\tt\footnotesize \href{mailto:fgaoaa@zju.edu.cn}{\color{black}fgaoaa}@zju.edu.cn}).}
\thanks{$^\dagger$ Corresponding authors: Shuai Wang and Yik-Chung Wu.
}
}

\markboth{IEEE Robotics and Automation Letters. Preprint Version. Accepted April, 2026}
{Zuo \MakeLowercase{\textit{et al.}}: DPNet: Doppler LiDAR Motion Planning for Highly-Dynamic Environments}

\maketitle

\begin{abstract}
Existing motion planning methods often struggle with rapid-motion obstacles due to an insufficient understanding of environmental changes. 
To address this, we propose integrating motion planners with Doppler LiDARs, which provide not only ranging measurements but also instantaneous point velocities. 
However, this integration is nontrivial due to the requirements of high accuracy and high frequency.
To this end, we introduce Doppler Planning Network (DPNet), which tracks and reacts to rapid obstacles via Doppler model-based learning. 
We first propose a Doppler Kalman neural network (D-KalmanNet) to track obstacle states under a partially observable Gaussian state space model.
We then leverage the predicted motions of obstacles to construct a Doppler-tuned model predictive control (DT-MPC) framework for ego-motion planning, enabling runtime auto-tuning of controller parameters.
These two modules allow DPNet to learn fast environmental changes from minimal data while remaining lightweight, achieving high frequency and high accuracy in both tracking and planning. 
Experiments on high-fidelity simulator and real-world datasets demonstrate the superiority of DPNet over extensive benchmark schemes.

\end{abstract}


\IEEEpeerreviewmaketitle

\vspace{-0.15in}
\section{Introduction}
\vspace{-0.05in}

\IEEEPARstart{R}{eal-time} robot motion planning in highly-dynamic environments is crucial for a broad range of applications, including emergency rescue and autonomous driving \cite{han2025hierarchically}.
Such scenarios require accurate knowledge of environmental dynamics, which can be leveraged by the motion controller to generate sequences of collision-free actions under kinematic constraints. 
Existing motion planning methods heavily rely on light detection and ranging (LiDAR) sensors to capture the current positions of the obstacles from the point cloud~\cite{flores2018cooperative}, and predict obstacles future trajectories via learning-based \cite{milan2017online}, physics-based \cite{lefkopoulos2020interaction}, or physics-informed learning \cite{revach2022kalmannet, liao2024physics} methods. 
Nonetheless, these approaches struggle to handle rapid-motion obstacles due to their limited ability to understand real-time environmental changes, such as instantaneous velocities.

Recently, however, the situation has changed thanks to the remarkable milestone of the emerging Doppler LiDAR sensors, which not only offer traditional point ranging data but also point Doppler velocity~\cite{hexsel2022dicp}.
The additional dimension explicitly provides instantaneous obstacle motion information, opening up new opportunities for motion planning in rapidly changing scenarios. 
Yet, how to integrate Doppler LiDAR into motion planners within a unified solution framework remains an open challenge.
This is nontrivial since the solution should be high-performance, real-time, safety-assured, and work properly with limited onboard computational resources.

\begin{figure}[!t] 
\centering
\includegraphics[width=0.47\textwidth]{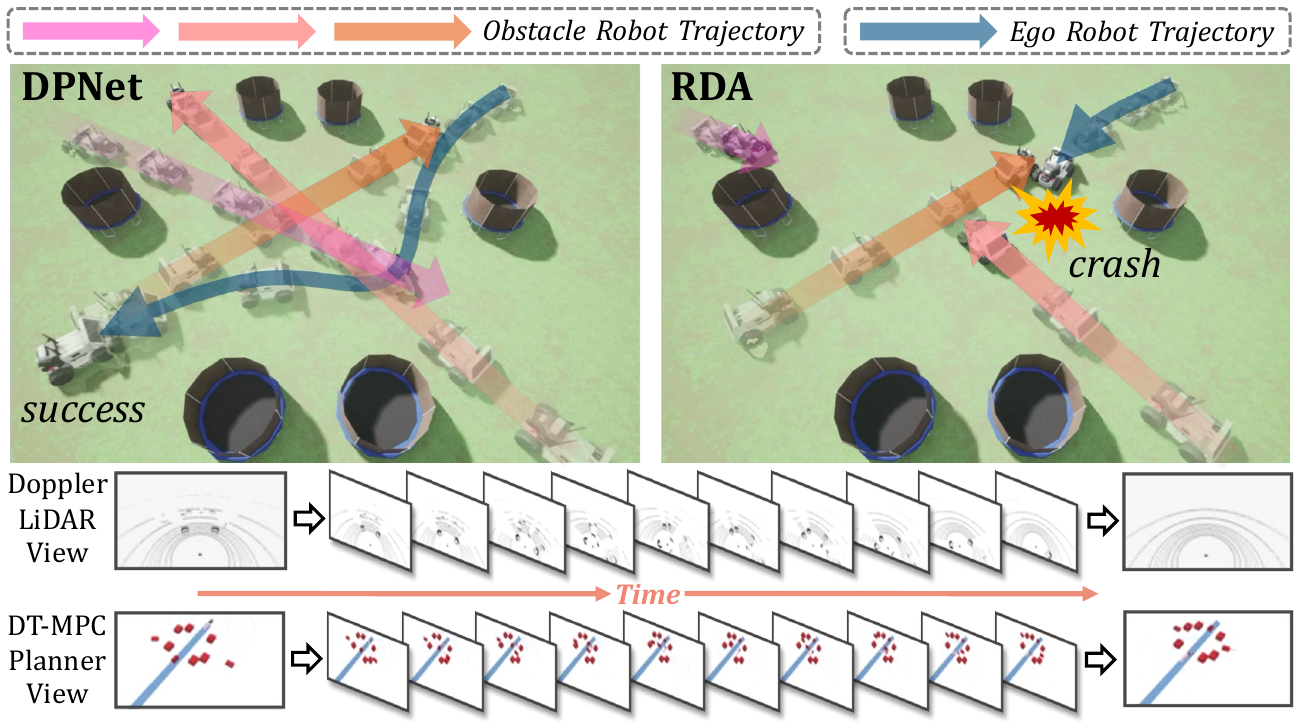}
\vspace{-0.02in}
\caption{Comparison between DPNet and RDA~\cite{han2023rda}. DPNet can understand and traverse highly-dynamic environments safely.}
\label{fig:overview}
\vspace{-0.2in}
\end{figure}

Given certain predictions of obstacle future states, the next step is motion control.
Conventional motion controllers, e.g., model predictive control (MPC), use the predicted states to construct consecutive problems along the horizon~\cite{zhang2024online}. However, they may suffer from suboptimality if the prior knowledge about the cost function or the dynamics model is imperfect. 
To address this, emerging auto-tuning MPC enables parameter learning from closed-loop execution data~\cite{tao2024difftune}.
Nevertheless, when applied to collision avoidance, these tune-from-execution methods typically require actual collision events to trigger the process.
To broaden their applications, how to leverage Doppler information to trigger runtime tuning without relying on actual events becomes an important issue. 

To fill these research gaps, this paper proposes Doppler planning network (DPNet). 
As shown in Fig.~\ref{fig:overview}, DPNet can traverse highly-dynamic areas by leveraging Doppler LiDAR and introducing two key algorithmic innovations for obstacle motion learning and runtime controller tuning. 
The first innovation is a Doppler Kalman neural network (D-KalmanNet) which achieves real-time obstacle future state prediction under a partially observable Gaussian state space (GSS) model. 
The D-KalmanNet incorporates the structural GSS model with a recurrent neural network (RNN) and sequences of Doppler velocity measurements, ensuring a very high tracking frequency and a lightweight model size. 
The second innovation is a Doppler-tuned model predictive control (DT-MPC) framework that performs Doppler-inferred collision check to trigger runtime parameter tuning from imagined collisions instead of actual ones. 
For computational efficiency, a heuristic update strategy is designed and an alternating direction method of multipliers (ADMM) algorithm is used for controller execution. 
Built upon these innovations, DPNet can understand and cope with rapid environmental changes using
minimal data.

To evaluate DPNet with D-KalmanNet and DT-MPC algorithms, extensive robot operation system (ROS) experiments are performed. 
We show that DPNet outperforms state-of-the-art planners including RDA \cite{han2023rda}, MPC-D-CBF \cite{jian2023dynamic}, and OBCA~\cite{zhang2020optimization} on diverse metrics in Carla simulator~\cite{dosovitskiy2017carla}.
% \textcolor{blue}{
Ablation study and sensitivity analysis confirm DT-MPC's effectiveness and robustness in triggering parameter updates without actual events, especially in cluttered environments.
% }
Moreover, evaluations on real-world Doppler LiDAR dataset Aevascene~\cite{aevascenes} demonstrate that D-KalmanNet achieves lowest tracking errors over existing tracking methods including Kalman filter \cite{jian2023dynamic} and KalmanNet \cite{revach2022kalmannet} across wide execution frequency range (i.e., $1–10$\,Hz), different horizon lengths ($1-10$), and various road scenarios (i.e., \textit{Highway} and \textit{City}).
It is also found that D-KalmanNet can be executed at $15-100$\,Hz on a low-cost NVIDIA Jetson Orin NX 16\,GB when tracking up to 10 objects simultaneously.
To the best of our knowledge, \emph{this is the first attempt to integrate Doppler LiDAR with motion planning}. 
Our contributions are summarized as follows:
\begin{itemize}
    \item We introduce Doppler LiDAR to support motion planner designs, resulting in a novel DPNet framework. By incorporating Doppler velocities for obstacle prediction, DPNet can proactively avoid highly-dynamic obstacles.    
    \item We propose D-KalmanNet for obstacle state tracking and prediction by model-based learning, achieving higher accuracy than existing methods while maintaining low computational cost and high frequency. 
    \item We propose DT-MPC for runtime controller tuning by triggering updates from Doppler-inferred collisions.
    \item We implement the complete DPNet framework for evaluation. 
    Extensive results show significant improvements over existing schemes in terms of diverse metrics.
\end{itemize}

\vspace{-0.15in}
\section{Related Work}
\vspace{-0.05in}

\begin{figure}[!t]
\centering
\includegraphics[width=0.48\textwidth]{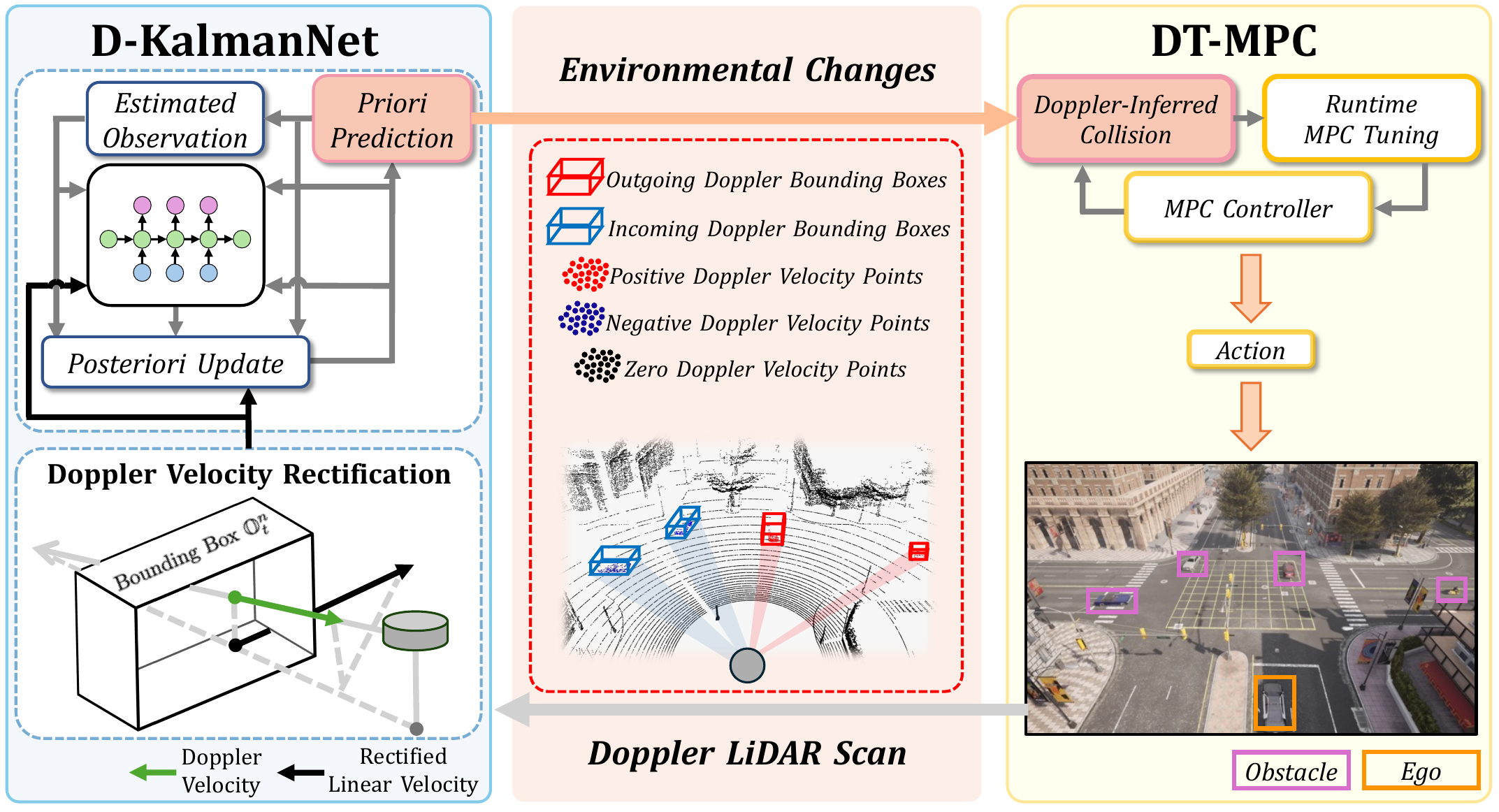}
\vspace{-0.02in}
\caption{The proposed DPNet system, which consists of D-KalmanNet and DT-MPC modules.}
\label{fig:system}
\vspace{-0.2in}
\end{figure}

\textbf{Doppler LiDAR.}
In contrast to traditional LiDAR~\cite{flores2018cooperative}, Doppler LiDAR reshapes the landscape of robotic solutions by providing Doppler velocity measurements. 
It has been used in object detection, state estimation, and simultaneous localization and mapping~\cite{shi2025pod,gu2022learning,peng2021detection,wu2022picking,zhao2024fmcw, yoon2023need, zhao2024free}
In terms of evaluation, Carla \cite{dosovitskiy2017carla} provides realistic Doppler LiDAR simulations~\cite{carla-aeva, hexsel2022dicp}, and AevaScenes~\cite{aevascenes} dataset provides on-road Doppler LiDAR recordings.
However, all existing research focuses on open-loop perception usages, and integrating Doppler LiDAR into closed-loop motion planning remains unexplored. 

\textbf{Obstacle Tracking.}
Incorporating forecasts of obstacle future trajectories can significantly enhance planning performance. 
Currently, obstacle tracking can be divided into learning-based, physics-based, and hybrid methods.
Learning-based methods require large datasets and substantial computational resources for training~\cite{zhou2025robust, zhang2022learning}.
Physics-based methods (e.g., Kalman filter \cite{lefkopoulos2020interaction}) capitalize on prior knowledge for high computational efficiency, while relying on predefined kinematic models that introduce inevitable bias and limit the ability to handle complex motions, especially when the state transition model is partially observed.
Lastly, hybrid methods combine strengths of both categories, resulting in efficient and adaptable physics-informed learning solutions.
For example, the recent KalmanNet \cite{revach2022kalmannet} mitigates the drawbacks of Kalman filters by learning the Kalman gain from real data, so as to maintain both high inference accuracy and lightweight model size. 
This work extends physics-informed learning to Doppler LiDAR, which has not been studied yet.

\textbf{Motion Planning.}
Many modern motion planning systems (e.g., robot, autonomous vehicle) are based on MPC, which generates optimal control outputs in response to environmental inputs within a receding horizon window \cite{shi2021advanced}.
By incorporating obstacle tracking, MPC can construct time-varying constraints and handle dynamic environments \cite{zhang2024online}. 
In general, the MPC performance depends on the problem modeling. 
To find the best setting, emerging auto-tuning MPC \cite{tao2024difftune} uses closed-loop feedback to optimize controller parameters. 
However, it typically involves time-consuming data collections or requires actual events to trigger.
In contrast, DT-MPC achieves runtime MPC tuning by inferring collisions from instantaneous Doppler measurements and obstacle state tracking, boosting real-time robustness for handling rapid-motion obstacles.

\section{Problem Formulation}\label{section2}

We consider an MPC motion planner operating among $N$ dynamic obstacles $\mathcal{N}=\{1,\cdots,N\}$.
At time $t$, the MPC horizon is represented by a finite sequence of $H$ discretized future steps
$\mathcal{H}_t\triangleq\{t,\cdots,t+H-1\}$, where $\Delta t$ is the time length between two states.
Let $\mathbf{s}_{h}=(x_h,y_h,\theta_h)$ be the robot state, where $(x_h,y_h)$ and $\theta_h$ are position and orientation at the bottom center of the corresponding bounding box $\mathbb{G}_h(\mathbf{s}_h)$, respectively. 
Let $\mathbf{w}_{h}=(\bm{v}_h,\psi_{h})$ be the robot action, where $\bm{v}_h$ and $\psi_{h}$ are linear and angular velocity, respectively.
By iterating over $\mathcal{H}_t$, the MPC system finds an optimal state sequence $\mathcal{S}_t=\{\mathbf{s}_{t+1},\mathbf{s}_{t+2},\cdots,\mathbf{s}_{t+H}\}$ via action sequence $\mathcal{W}_t=\{\mathbf{w}_t,\mathbf{w}_{t+1},\cdots,\mathbf{w}_{t+H-1}\}$, 
where $\{\mathcal{W}_t,\mathcal{S}_t\}\in\mathcal{F}_t$, with $\mathcal{F}_t$ being the kinematic constraint detailed in Section IV.
The problem is thus formulated as
\begin{subequations}
\label{main}
\begin{align}
    & \mathsf{P}_t:~\min_{\{\mathcal{W}_t,\mathcal{S}_t\}\in\mathcal{F}_t}\ C_t(\mathcal{S}_t)  \\
    & \text{s.t.}~~~~\mathbf{dist}(\mathbb{G}_{h+1},{\mathbb{O}}_{h+1}^n)\geq d_{\text{safe}}, \, \forall h \in \mathcal{H}_t, \, n\in\mathcal{N},
\end{align}
\end{subequations}
where $C_t(\mathcal{S}_t)$ is the utility function, $d_{\text{safe}}$ is the safety distance, and $(\mathbb{G}_{h+1},{\mathbb{O}}_{h+1}^n)$ are bounding boxes of ego-robot and the $n$-th obstacle, respectively.
The minimum distance function between two bounding boxes is defined as $\mathbf{dist}(\mathbb{P},\mathbb{Q})=\min\{\|\mathbf{p}-\mathbf{q}\|_2\, | \, \mathbf{p}\in\mathbb{P}, \mathbf{q}\in \mathbb{Q}\}$.
Considering a reference path tracking task with target waypoints $\{\mathbf{s}_{h+1}^{\star}|h \in \mathcal{H}_t\}$, we formulate $C_t(\mathcal{S}_t) = \sum_{h \in \mathcal{H}_t} \left\| \mathbf{s}_{h+1} - \mathbf{s}_{h+1}^{\star} \right\|_2^2$ \cite{zhang2020optimization}. 

Problem $\mathsf{P}_t$ is nontrivial since estimating ${\mathbb{O}}_{h+1}^n$ results in an inevitably biased  $\hat{\mathbb{O}}_{h+1|t}^n$, which can significantly mismatch the actual ${\mathbb{O}}_{h+1}^n$ if obstacle $n$ has rapid motions. 
This error would further propagate to motion planning since 
$\mathbf{dist}(\mathbb{G}_{h+1},\hat{\mathbb{O}}_{h+1|t}^n)
\neq
\mathbf{dist}(\mathbb{G}_{h+1},{\mathbb{O}}_{h+1}^n)
$.

\begin{figure}[!t]
\vspace{-0.1in}
\centering
\includegraphics[width=0.4\textwidth]{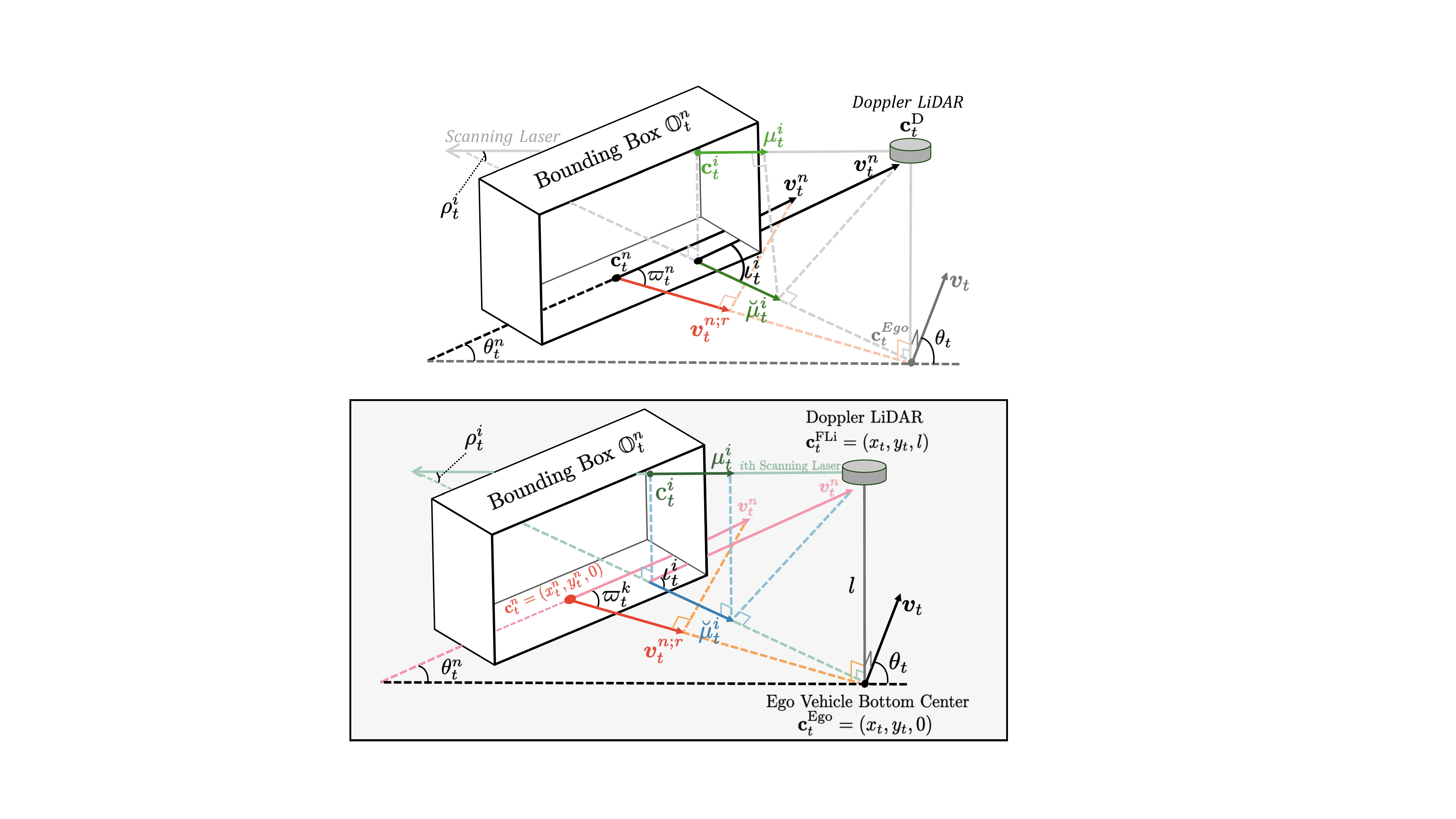}
\vspace{-0.05in}
\caption{Doppler velocity rectification.}
\label{fig:doppler}
\vspace{-0.15in}
\end{figure}

\vspace{-0.1in}
\section{Doppler Planning Network}\label{section3}

To solve $\mathsf{P}_t$ among rapid-motion obstacles, we propose D-KalmanNet for reducing the uncertainty of $\hat{\mathbb{O}}_{h+1|t}^n$ and DT-MPC for robustifing the solution by real-time tuning $d_{\text{safe}}$.  
We then integrate them into a unified solution, DPNet, whose overall architecture is shown in Fig. \ref{fig:system}. 
It takes the Doppler LiDAR point cloud $\mathcal{P}_t$ as input and outputs the actions $\mathcal{W}_t$.

\vspace{-0.15in}
\subsection{Doppler KalmanNet for Tracking}\label{AA}

The ego robot uses a Doppler LiDAR to scan the surrounding environment and obtain Doppler point cloud $\mathcal{P}_t=\{\mathbf{c}_t^i, \|\bm{\mu}_t^i\|_2\}_{i=1}^{I}$, where $I$ is the number of points in the frame, $\mathbf{c}^i_t\in \mathbb{R}^3$ and $\|\bm{\mu}^i_t\|_2 \in \mathbb{R}$ are the spatial coordinate and the scalar Doppler velocity of the $i$-th point, respectively. 
Accordingly, we define the Doppler-augmented state $\mathbf{x}_t^n$ as
    \begin{align}
        \mathbf{x}^n_t=[
        x_t^n,\cos\theta_t^n\bm{v}_t^n,\cos\theta_t^n\bm{a}_t^n, 
        y_t^n, \sin\theta_t^n\bm{v}_t^n,\sin\theta_t^n\bm{a}_t^n
    ]^\top, \label{xtn}
    \end{align}
where $(x_t^n,y_t^n)$, $\theta_t^n$, $\bm{v}_t^n$ and $\bm{a}_t^n$ represent the position, orientation, linear velocity and acceleration at the center, respectively.
Each $\mathbf{x}_t^n$ can be used to generate a ${\mathbb{O}}_{t}^n$. 
Hence, we only need to estimate a sequence of 
$\{\hat{\mathbf{x}}_{t+1|t}^n, \hat{\mathbf{x}}_{t+2|t}^n, \ldots, \hat{\mathbf{x}}_{t+H|t}^n\}$ from $\mathcal{P}_t$.
We can fuse $\mathcal{P}_t$ with a state transition model to achieve this goal, but two challenges exist. 
First, according to Doppler LiDAR manufacturers \cite{aevascenes}, real-world velocity measurements typically involve noises with standard deviation of $0.1$\,m/s.
Second, built upon past trajectories, the transition model may mismatch the actual obstacle motions in the future \cite{revach2022kalmannet}. 
Our D-KalmanNet can simultaneously tackle both challenges.

\subsubsection{Doppler Velocity Rectification}
We propose Doppler velocity rectification shown in Fig. \ref{fig:doppler} to address the first challenge.  
Our observation is that any measurement point on a common obstacle with a rigid-body motion would share the same linear velocity.
As such, it is possible to aggregate velocities across dense points for mitigating the Doppler LiDAR noise.
Since different lasers hit different points on obstacle surfaces and reflect different radial velocities, we first unify the measurements to the same vector direction. 
Specifically, let $\mathbf{c}_t^n, \mathbf{c}^\text{Ego}_t\in\mathbb{R}^3$ represent positions of obstacle bottom center and ego robot bottom center, respectively. 
The Doppler LiDAR mounted at $\mathbf{c}^\text{D}_t\in\mathbb{R}^3$ emits its $i$-th scanning laser with a scanning elevation angle $\rho_t^i$, hitting and reflecting from point $\mathbf{c}_t^i$. 
The Doppler velocity projection $\bm{\breve{\mu}}_t^i$ on the 2D plane is derived as:
    \begin{align}
    \bm{\breve{\mu}}_t^i=\frac{\|\bm{\mu}_t^i\|_2}{\cos\rho_t^i}\times\frac{(\mathbf{I}-\mathbf{e}\mathbf{e}^\top)(\mathbf{c}^{\text{D}}_t-\mathbf{c}_t^i)}{\|(\mathbf{I}-\mathbf{e}\mathbf{e}^\top)(\mathbf{c}^{\text{D}}_t-\mathbf{c}_t^i)\|_2},
    \label{radial_solve}
    \end{align}
where $\mathbf{e}=\begin{bmatrix}0 & 0 & 1\end{bmatrix}^\top$ and $\mathbf{I}\in\mathbb{R}^{3\times3}$ is an identity matrix.

\begin{algorithm}[t]
\caption{D-KalmanNet}
\label{algorithm1}
\textbf{Input}: $\{\mathbf{y}_0^n, \mathbf{y}_1^n,\cdots\}_{n=1}^N$\\
\textbf{Output}: $\{\hat{\mathcal{O}}_0^n, \hat{\mathcal{O}}_1^n, \cdots\}_{n=1}^N$
\begin{algorithmic}[1]
\STATE Initialize $t=0$
\WHILE{\textit{True}}
   \FOR {$n\in\mathcal{N}$}
    \STATE Group $\mathcal{P}_t^n$ to obtain $\mathbf{y}_t^n$
    \STATE \textbf{if} $t=0$ \textbf{then} ~~\tcp{\small Initial Guess}
       \STATE \quad $\mathbf{x}_0^n \gets \mathbf{y}_0^n$, $\hat{\mathbf{x}}_{1|0}^n \gets \mathbf{T}\cdot \mathbf{x}_0^n$, $\hat{\mathbf{y}}_{1|0}^n \gets \mathbf{U} \cdot \hat{\mathbf{x}}_{1|0}^n$ 
    \STATE \textbf{else} ~~\tcp{\small Kalman Learning}
        \STATE \quad $\mathcal{K}_{t}^n \gets \textsf{RNN}(\mathbf{x}^n_{t-1},\hat{\mathbf{x}}_{t|t-1}^n, \hat{\mathbf{y}}_{t|t-1}^n, \mathbf{y}_t^n)$
        \STATE \quad $\mathbf{x}_{t}^n \gets \hat{\mathbf{x}}_{t|t-1}^n + \mathcal{K}_{t}^n\cdot(\mathbf{y}_{t}^n - \hat{\mathbf{y}}_{t|t-1}^n)$
        \STATE \quad  $\hat{\mathbf{x}}_{h+1|t}^n \gets \mathbf{T}^{h-t+1} \cdot \mathbf{x}_t^n$, $\forall h \in \mathcal{H}_t $
        \STATE \quad $\hat{\mathbf{y}}_{t+1|t}^n \gets \mathbf{U} \cdot \hat{\mathbf{x}}_{t+1|t}^n$ 
    \STATE \textbf{end if}
    \STATE return $\hat{\mathcal{O}}_{t}^n=\{\hat{\mathbb{O}}_{h+1|t}^n(\hat{\mathbf{x}}_{h+1|t}^n)\}_{\forall h\in\mathcal{H}_t}$
   \ENDFOR  
        \STATE $t \gets t+1$
\ENDWHILE
\end{algorithmic}
\end{algorithm}

With the projection, we then group Doppler points inside the obstacle bounding box as $\mathcal{P}_t^n=\{\{\mathbf{c}_t^i, \|\bm{\mu}_t^i\|_2\}|\mathbf{c}_t^i\sqsubseteq \mathbb{O}^n_t\}=\{p_t^{i_n}\}_{i_n=1}^{I_n}$, where $p_t^{i_n}$ is the $i_n$-th point of group $\mathcal{P}_t^n$, $\sqsubseteq$ is an operator that indicates being within the spatial boundary. 
Denote $\varpi_t^n$ as the angle between $\bm{v}_t^n$ and center radial velocity $\bm{v}_t^{n;r}$, and $\iota_t^i$ as the angle between $\bm{v}_t^n$ and Doppler velocity projection $\bm{\breve{\mu}}_t^i$. 
Based on rigid motion consistency $\bm{v}_t^{n;r}\cos\iota_t^i=\bm{\breve{\mu}}_t^i \cos\varpi_t^n$, the fused linear velocity $\bm{v}_t^n$ is given by
\vspace{-0.05in}
\begin{align}
        \bm{v}_t^n=\frac{1}{|I_n|}\sum_{i_n=1}^{I_n}\frac{1}{\cos\iota_t^{i_n}}\bm{\breve{\mu}}_t^{i_n}. \label{fused_vt}
    \end{align}
\vspace{-0.1in}    

\subsubsection{Kalman Gain Learning}

We propose Kalman gain learning to address the second challenge. 
We set up a partially observable GSS model for the dynamics, whose state transition model is $\mathbf{T}$ (assuming a constant acceleration) and state observation model is ${\mathbf{U}}$ (involving both position and velocity). 
Then we have the following state transition
    \begin{align}
        \hat{\mathbf{x}}^n_{h+1|h}=\mathbf{T}\mathbf{x}^n_h, \ \hat{\mathbf{y}}^n_{h+1|h}={\mathbf{U}}\hat{\mathbf{x}}^n_{h+1|h},
    \end{align}    
where $\hat{\mathbf{x}}^n_{h+1|h}$ is the priori state prediction and $\hat{\mathbf{y}}^n_{h+1|h}$ is the estimated state observation using forward propagation.
After $\Delta t$, new scan $\mathcal{P}_{h+1}$ arrives, providing the actual $\mathbf{y}^n_{h+1}$ by \eqref{xtn} and \eqref{fused_vt}. 
We then update the posteriori state prediction by
\vspace{-0.05in}
\begin{align}
    \mathbf{x}_{h+1}^n&=\hat{\mathbf{x}}_{h+1|h}^n+\mathcal{K}_{h+1}^n\cdot(\mathbf{y}^n_{h+1}-\hat{\mathbf{y}}^n_{h+1|h}),
\end{align}
where $\mathcal{K}_{h+1}^n$ is the learned Kalman gain:
\begin{align}
    &\mathcal{K}_{h+1}^n =\textsf{RNN}(\mathbf{x}^n_{h},\hat{\mathbf{x}}_{h+1|h}^n, \hat{\mathbf{y}}_{h+1|h}^n, \mathbf{y}_{h+1}^n).
    \label{eq:update}    
\end{align}

We adopt an RNN \cite{revach2022kalmannet} to learn a robust and adapative $\mathcal{K}_{h+1}^n$ in \eqref{eq:update} for posteriori update, thereby alleviating the noises originated from either mismatched $\mathbf{T}$ or inaccurate ${\mathbf{y}}_{h+1|h}^n$.
The entire procedure is summarized in Algorithm 1, where $\hat{\mathcal{O}}_{t}^n=\{\hat{\mathbb{O}}_{h+1|t}^n(\hat{\mathbf{x}}_{h+1|t}^n)\}_{\forall h\in\mathcal{H}_t}$ represents bounding box predictions over the planning horizon.

\begin{algorithm}[!t]
\caption{Doppler Collision Check-and-Tune}
\label{algorithm2}
\begin{algorithmic}[1] %[1] enables line numbers
\STATE Initialize $\mathcal{S}_{t-1}=\emptyset$
\WHILE{\textit{True}}
\STATE $\varrho_n \gets \infty$ for $n\in\mathcal{N}$, $\kappa \gets \kappa^\text{init}$ ~~\tcp{\small New Iteration}
\STATE Obtain $\hat{\mathcal{O}}_{t}^n$\ from the D-KalmanNet
\FOR {$h\in\mathcal{H}_{t-1}^-$}
\FOR {$n\in\mathcal{N}$}
\STATE \textbf{if} {$\varrho_n = \infty$ \textbf {and} $\mathbf{dist}(\mathbb{G}^\circ_{h+1},\hat{\mathbb{O}}_{h+1|t}^n) \leq d_0$} \textbf{then}
\STATE \quad $\varrho_n \gets \kappa$  ~~\tcp{\small Assign Collision Priority}
\STATE \textbf{end if}
\ENDFOR
\STATE Update $\kappa \gets \kappa + \Delta\kappa$ ~~\tcp{\small Update Accumulator}
\ENDFOR 
\STATE $t \gets t+1$
\ENDWHILE
\end{algorithmic}
\end{algorithm}

\vspace{-0.1in}
\subsection{Doppler-Tuned MPC for Planning}

A fixed $d_{\text{safe}}$ in $\mathsf{P}_t$ often leads to poor performance or problem infeasibility~\cite{tao2024difftune}.
This can be addressed by MPC auto-tuning~\cite{han2025neupan}, which transforms $d_{\text{safe}}$ into a learnable parameter vector 
$\bm{\phi}=[\phi_{t+1,1},\cdots,\phi_{t+H,N}]^T$ and reformulates $\mathsf{P}_t$ into the following parameter optimization problem:
\begin{subequations}
    \begin{align} 
    &\mathsf{Q}_t:\underset{\{\mathcal{W}_t,\mathcal{S}_t\}\in\mathcal{F}_t,\boldsymbol{\phi} \in \Phi}{\text{min}} \  C_t(\mathcal{S}_t)+\gamma\mathcal{L}(\boldsymbol{\phi},\{\hat{\mathcal{O}}_t^n\}),
\end{align}
\end{subequations}
where $\mathcal{L}$ is the closed-loop collision cost function, $\gamma$ is the collision penalty coefficient and $\Phi$ defines the upper and lower bounds of $\bm{\phi}$.
One example of $\mathcal{L}$ is the quadratic loss of the collision penalty \cite{han2025neupan}, where 
\vspace{-0.05in}
\begin{align}
&\mathcal{L}(\boldsymbol{\phi},\{\hat{\mathcal{O}}_t^n\})=
\nonumber\\
&
    \sum_{h\in\mathcal{H}_t}
    \sum_{n\in\mathcal{N}}
\Big\|\mathrm{min} \Big(\mathbf{dist}(\mathbb{G}_{h+1},\hat{\mathbb{O}}_{h+1|t}^n)
-\phi_{h+1,n},0\Big)
\Big\|^2.
\end{align}
\vspace{-0.1in}

To minimize $\mathcal{L}$,
DT-MPC adopts a heuristic rule~\cite{zhang2024multi} to trigger parameter updates by using $\hat{\mathcal{O}}^n_t$ for Doppler-inferred collision check. This allows exploring whether real-time Doppler measurements can facilitate tuning even with simple heuristics (see Section V-D for its limitation).

\subsubsection{Doppler-Inferred Collision Check}

In an MPC system, solving $\mathsf{Q}_t$ produces complete $\mathcal{W}_t$ and $\mathcal{S}_t$ along $\mathcal{H}_t$, while only the first action $\mathbf{w}_t$ is executed, which ends up with state $\mathbf{s}_{t+1}$. This leaves the potential state sequence $\mathcal{S}_t^\circ=\{\mathbf{s}_h^\circ\}_{h=t+2}^{t+H}$ as a natural surrogate of ego trajectory prediction.
Hence, before solving $\mathsf{Q}_t$, we perform parameter tuning by assessing historical potential state sequence $\mathcal{S}_{t-1}^\circ$ from the solution of $\mathsf{Q}_{t-1}$.
Specifically, $\mathcal{S}_{t-1}^\circ$ corresponds to the ego bounding box prediction $\{\mathbb{G}^\circ_{h+1}(\mathbf{s}_{h+1}^\circ)\}_{\forall h\in\mathcal{H}_{t-1}^-}$, where $\mathcal{H}_{t-1}^-=\mathcal{H}_{t-1}\setminus \{t-1\}$ denotes the truncated historical horizon.
% \textcolor{blue}{
Inspired by~\cite{schulman2014motion}, we then check potential collisions between $\{\hat{\mathbb{O}}_{h+1|t}^n(\hat{\mathbf{x}}_{h+1|t}^n)\}$ and $\{\mathbb{G}^\circ_{h+1}(\mathbf{s}_{h+1}^\circ)\}$ along ${\mathcal{H}_{t-1}^-}$ , using the following criteria:
% }
    \begin{align}
        \mathbf{dist}(\mathbb{G}^\circ_{h+1},\hat{\mathbb{O}}_{h+1|t}^n) \leq d_0, \forall h\in\mathcal{H}^{-}_{t-1},n\in\mathcal{N}, \label{d-check}
    \end{align}
where $d_0$ is the distance threshold of Doppler-inferred collision. Note that the last element of $\hat{\mathcal{O}}_t^n$ is excluded from the check to match the length of $\mathcal{H}_{t-1}^-$.

\begin{figure*}[!t]
\centering
\includegraphics[width=0.98\textwidth]{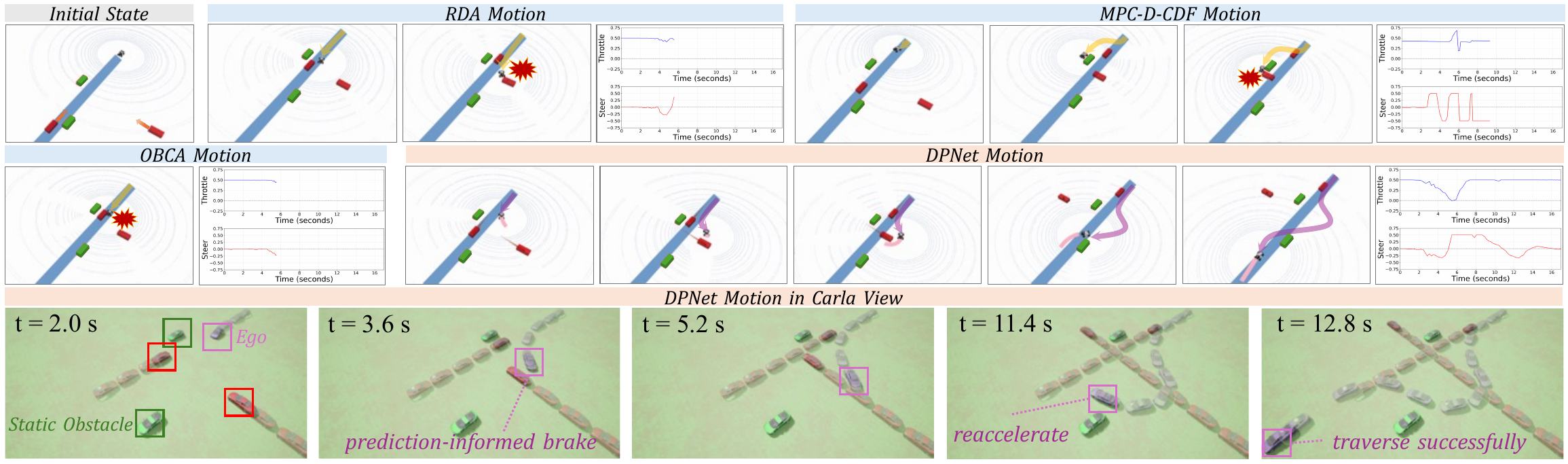}
\vspace{-0.05in}
\caption{Qualitative analysis of robot motions (in ROS-RViz and Carla~\cite{dosovitskiy2017carla} views) and corresponding control commands. Static obstacles are marked as green boxes, while dynamic ones are red boxes with arrows indicating their moving directions. The blue line is the path connecting start and goal points. The dots in front of dynamic obstacles represent motion predictions.}
\label{fig:motion}
\vspace{-0.2in}
\end{figure*}

\subsubsection{Run-Time Auto-Tuning}
If such Doppler-inferred collision \eqref{d-check} exists for any obstacle, $\bm{\phi}$ for $\mathsf{Q}_{t}$ is updated to alleviate potential future risks.
We first decompose $\phi_{h,n}$ into a constant component and an adaptive component as $\phi_{h,n}=d_{1}+\tau(n,h) d_2$, where $d_1$ is the compulsory minimum distance and $\tau(n,h) d_2$ is the Doppler adaptive distance. 
This converts tuning $\phi_{h,n}$ to tuning the factor $\tau(n,h)$.
We further decompose $\tau(n,h)=\tau_1(n)\tau_2(h)$, where $\tau_1(n) \in (0,1)$ is the spatial factor and $\tau_2(h) \in (0,1)$ is the temporal factor. First, the spatial factor $\tau_1(n)$ measures the imminence of the Doppler-inferred collision by
\vspace{-0.05in}
\begin{align}
        \tau_1(n) = \max\left(1 - \alpha (\varrho_n - \kappa^\text{init}), \tau_{1;\min}\right),
\end{align}
where $\tau_{1;\min}$ is the lower bound, $\alpha\in(0,1)$ is the decay rate, and $\varrho_n$ is the collision priority assigned chronologically. A larger $\alpha$ leads to a higher focus on imminent collisions.
Algorithm \ref{algorithm2} details how $\varrho_n$ is assigned using the accumulator $\kappa$, which starts with $\kappa^\text{init}$ and increments by $\Delta \kappa$ along the horizon.
Second, the temporal factor $\tau_2(h)$ accounts for the inherently increasing uncertainty of farther predictions:
\vspace{-0.05in}
\begin{align}
        \tau_2(h) = \max\left(1 - \beta (h - t), \tau_{2;\min}\right),
\end{align}
where $\tau_{2;\min}$ is the lower bound and $\beta\in(0,1)$ is the horizon decay rate. A larger $\beta$ imposes stricter constraints on nearby predictions while relaxing constraints for distant ones. Table~\ref{tab:sens} in Section V provides a sensitivity analysis of $(\alpha,\beta)$.

\subsubsection{Solution by ADMM}
With the Doppler-tuned $\hat{\boldsymbol{\phi}}$ from DT-MPC (Algorithm 2), $\mathsf{Q}_t$ is explicitly written as
\vspace{-0.02in}
\begin{subequations}
\label{main2}
\begin{align}
    & \mathsf{R}_t:~\min_{\mathcal{W}_t,\mathcal{S}_t}\ C_t(\mathcal{S}_t)+\gamma\mathcal{L}(\hat{\boldsymbol{\phi}},\{\hat{\mathcal{O}}_t^n\}),  \\
    & \text{s.t.}~~~~\mathbf{s}_{h+1} = \mathbf{s}_{h} + f(\mathbf{s}_{h}, \mathbf{w}_{h}) \Delta t, \ \forall h, \label{sta_tran}  \\
    & ~~~~~~~\mathbf{w}_{\min} \preceq \mathbf{w}_{h} \preceq \mathbf{w}_{\max}, \ \forall h, \label{limit}
\end{align}
\end{subequations}
where \eqref{sta_tran} represents the state evolution ($f(\cdot)$ defines vehicle dynamics)
and \eqref{limit} represents the physical constraint ($\mathbf{w}_{\min},\mathbf{w}_{\max}$ are control limits),
i.e., the kinematic set $\mathcal{F}_t$ explicitly becomes $\mathcal{F}_t=\{\eqref{sta_tran}, \eqref{limit}\}$ \cite{zhang2020optimization}. 
To tackle the only nonconvex part $\mathcal{L}(\hat{\boldsymbol{\phi}},\{\hat{\mathcal{O}}_t^n\})$ in $\mathsf{R}_t$, we equivalently transform $\mathcal{L}$ into a biconvex form using Lagrange duality \cite{han2023rda,han2025neupan}, and then solve the resultant biconvex version by ADMM. This ADMM will converge to a stationary point of $\mathsf{R}_t$.

\section{Experiments}\label{section5}

We implement DPNet in ROS Noetic on Linux system, using Carla simulator with Doppler LiDAR integration \cite{carla-aeva}.
We compare DPNet with state-of-the-art methods and provide both qualitative and quantitative results in highly-dynamic environments randomize by DynaBARN~\cite{nair2022dynabarn} benchmark, along with ablation study and sensitivity analysis for DT-MPC.
We also evaluate D-KalmanNet against existing obstacle tracking methods in highly-dynamic environments, and assess its efficiency on resource-constrained platform NVIDIA Jetson Orin NX.
Ground truths of obstacle bounding boxes are used to group Doppler LiDAR points.

We train D-KalmanNet for 2000 epochs on the real-world Doppler LiDAR dataset AevaScenes~\cite{aevascenes} comprising 100 sequences (50 in \textit{City}, 50 in \textit{Highway}). Each sequence contains 100 frames recorded at 10\,Hz with ground-truth vehicle linear velocities and bounding box poses. Training uses 80 randomly selected sequences (40 per scenario), and the remaining 20 sequences serve as the evaluation set in Section V-B.
 
\vspace{-0.1in}
\subsection{End-to-End Evaluation in Carla}

We compare the following methods\footnote{Reinforcement learning baselines are not included, since they suffer from a lack of safety assurance in highly-dynamic scenarios.} for end-to-end evaluations in Carla-ROS simulation:
\textbf{(1) DPNet}: our proposed method;
\textbf{(2) MPC-D-CBF} \cite{jian2023dynamic}: an MPC-based planner that predicts obstacle motion using LiDAR and Kalman filter, constructing D-CBFs online for navigation among moving obstacles;
\textbf{(3) RDA} \cite{han2023rda}: an MPC-based planner that uses parallelizable dual ADMM to accelerate collision avoidance;
\textbf{(4) OBCA} \cite{zhang2020optimization}: an MPC-based planner with optimization-based collision avoidance for convex obstacles; 
\textbf{(5) DPNet (Ablation)}: our proposed method where DT-MPC is disabled.
All methods are set up with $H=15$ and $\Delta t=0.1\,$s.
DPNet's specific parameters are set as follows:
$d_0=0.5\,$m, $d_1=0.1\,$m, $d_2=0.4\,$m, $\alpha=0.2$, $\beta=0.05$, $\tau_{1;\min}=0.3$, $\tau_{2;\min}=0.3$, $\kappa^\text{init}=1.0$, and $\Delta\kappa=1.0$. 
Ego robot and obstacles are modeled as car-like wheel robots.

\textbf{Qualitative Analysis.}
As shown in Fig. \ref{fig:motion}, this scenario consists of two static obstacles and two rapid-motion obstacles (at $5\,$m/s) with frontal and lateral directions. 
These obstacles \emph{will not avoid the ego robot}. 
The ego robot needs to traverse the area without any collisions. 
It can be seen that our DPNet succeeds to pass through the area with agile motions. 
Particularly, at $t=2\,$s, if velocity information is ignored, directly passing between the two dynamic objects appears to be the optimal choice. 
However, at $t=4\,$s, the ego-robot is completely surrounded by four obstacles, and it was too late to steer by then.
This is case of RDA, which crashes into the lateral moving obstacle at $t=5.6\,$s.
In contrast, \emph{with accurate velocity understanding, our DPNet can look ahead into the future, ``imagining'' that directly passing through is risky, and thus early-executes the left-turn at $t=2\,$s}. 
As such, at $t=3.6\,$s and $t=5.2\,$s, the ego-robot bypasses from behind the laterally moving obstacle.
This demonstrates the power of Doppler model-based learning in handling highly-dynamic environments. 
In terms of other methods, OBCA crashes into the frontally moving obstacle due to its delayed motions caused by low computation frequency. MPC-D-CBF planner makes over-conservative decision upon handling the frontally moving obstacle due to its inaccurate velocity understandings, resulting in a crash eventually.

\textbf{Quantitative Analysis.}
To evaluate DPNet in highly-dynamic environments, we adopt DynaBARN~\cite{nair2022dynabarn} benchmark to set up obstacles with randomized trajectories, velocities ($6\pm2\,$m/s), and accelerations.
The following metrics are used: (a) \textbf{AvgAcc\,(m/s\(^2\))}: the average acceleration during the execution; (b) \textbf{MaxAcc\,(m/s\(^2\))}: the maximum acceleration duing the execution; (c) \textbf{AvgJerk\,(m/s\(^3\))}: the average jerk during the execution; (d) \textbf{IteTime\,(ms)}: the average iteration time of the solver during the execution; (e) \textbf{PassTime\,(s)}: the average time duration to traverse the area; (f) \textbf{PassRate\,(\%)}: the rate of successfully traversing the area without collisions.

\begin{figure}[t] 
\centering
\includegraphics[width=0.45\textwidth]{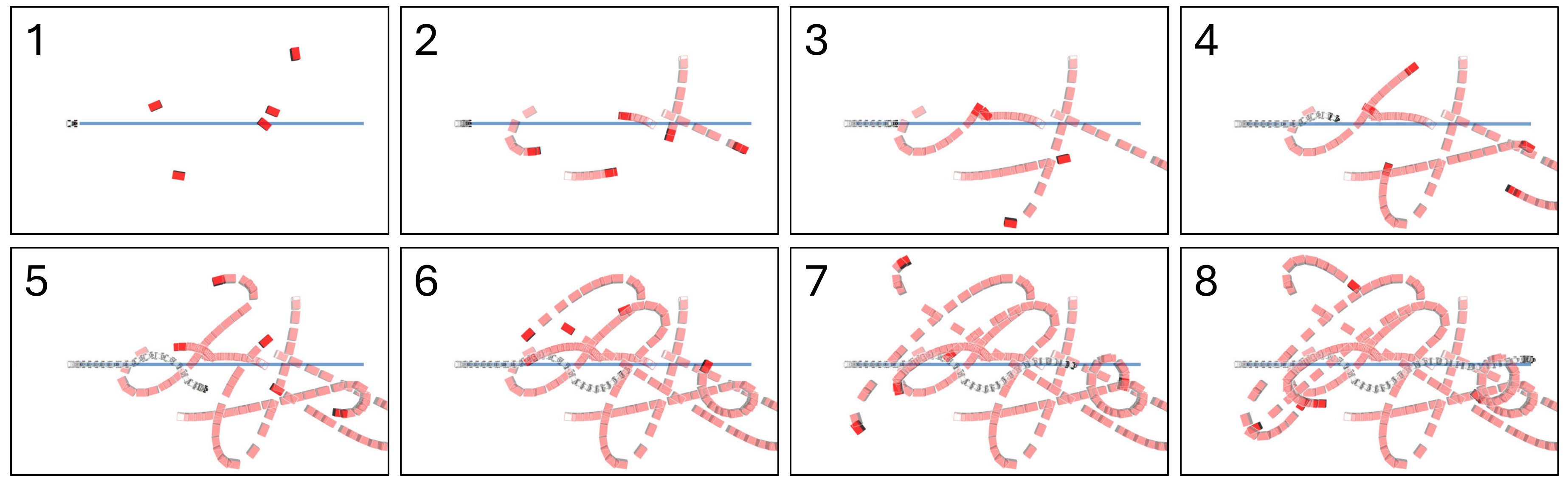}
\caption{Visualization of DPNet traversing a  highly-dynamic 5-obstacle area,  randomized by DynaBARN~\cite{nair2022dynabarn} benchmark. Frames are sampled by $1.25$ second interval. 
}
\label{fig:random}
\vspace{-0.21in}
\end{figure}

Results are reported in Table \ref{tab:performance}, where each obstacle setting is executed 100 times. 
As the number of obstacles varies from 1 to 7, DPNet consistently outperforms other baselines, with a significantly superior PassRate in dense obstacle settings. 
In terms of computational efficiency, DPNet achieves a low IteTime, close to that of RDA~\cite{han2023rda}.
However, PassRate of RDA is far below that of DPNet, e.g., degrading by $42.2\,\%$ when the number of obstacles is 7.
These results confirm that DPNet robustifies collision avoidance with efficient computation.
Ablation study shows that without DT-MPC, DPNet's PassRate drops especially under high obstacle numbers, demonstrating the effectiveness of leveraging real-time Doppler measurements for MPC tuning.
Fig. \ref{fig:random} demonstrates how DPNet traverses an area with 5 randomized highly-dynamic obstacles.

Table~\ref{tab:sens} presents a sensitivity analysis of DT-MPC under different parameter settings of ($\alpha$, $\beta$), with 100 runs for each setting. 
DT-MPC maintains stable performance across various settings. 
Increasing $\alpha$ or $\beta$ slightly improves PassRate while degrading AvgJerk, since larger $\alpha$ or $\beta$ leads to more conservative actions (e.g., abrupt turns) for closer obstacles.
Note that DT-MPC achieves a low tuning time $t_{\text{DT-MPC}}$ ($<1.5\,$ms), demonstrating its real-time efficiency.

\begin{table}[t]
\centering
\caption{Quantitative Comparison}
\vspace{-0.02in}
\begin{tabular}{cccccc}
\toprule

\multirow{3}{*}{\textbf{Metric}}&\multicolumn{5}{c}{\textbf{Method}}\\
\cmidrule{2-6}
&{OBCA}&{RDA}&{MPC-D-CBF}&DPNet&\multirow{2}{*}{DPNet} \\
&\cite{zhang2020optimization}&\cite{han2023rda}&\cite{jian2023dynamic}&(Ablation)&\\

\midrule
-&\multicolumn{5}{c}{1 obstacle}\\
\addlinespace[1pt]
AvgAcc\,$\downarrow$   &1.628 &\cellcolor{purple2}1.425 &1.725 &1.496 &\cellcolor{purple1}1.414\\
MaxAcc\,$\downarrow$   &8.415 &8.596 &8.506 &\cellcolor{purple2}7.905 &\cellcolor{purple1}7.836 \\
AvgJerk\,$\downarrow$  &7.996 &7.004 &7.671 &\cellcolor{purple1}6.873 &\cellcolor{purple2}6.965 \\
IteTime\,$\downarrow$  &147.5 &88.0\cellcolor{purple1} &128.3 &99.4\cellcolor{purple2} &102.1 \\
PassTime\,$\downarrow$ &10.362 &8.852 &9.446 &\cellcolor{purple1}7.976 &\cellcolor{purple2}8.020 \\
PassRate\,$\uparrow$   &76.0 &87.0 &\cellcolor{purple1}100.0 &\cellcolor{purple2}99.0 &\cellcolor{purple1}100.0 \\

\midrule
-&\multicolumn{5}{c}{3 obstacles}\\
AvgAcc\,$\downarrow$   &1.743 &1.599 &2.139 &\cellcolor{purple1}1.537 &\cellcolor{purple2}1.582 \\
MaxAcc\,$\downarrow$   &10.085 &9.622 &9.878 &\cellcolor{purple2}9.561 &\cellcolor{purple1}9.284 \\
AvgJerk\,$\downarrow$  &9.251 &\cellcolor{purple2}8.096 &10.032 &8.101 &\cellcolor{purple1}8.058 \\
IteTime\,$\downarrow$  &163.8 &\cellcolor{purple1}114.5 &151.7 &137.7 &\cellcolor{purple2}135.3 \\
PassTime\,$\downarrow$ &12.550 &11.376 &14.238 &\cellcolor{purple2}10.497 &\cellcolor{purple1}10.194 \\
PassRate\,$\uparrow$   &41.0 &54.0 &74.0 &\cellcolor{purple2}85.0 &\cellcolor{purple1}88.0 \\

\midrule
-&\multicolumn{5}{c}{5 obstacles}\\
AvgAcc\,$\downarrow$   &1.869 &1.606 &2.213 &\cellcolor{purple2}1.598 &\cellcolor{purple1}1.571 \\
MaxAcc\,$\downarrow$   &9.515 &\cellcolor{purple2}9.481 &9.626 &\cellcolor{purple1}9.349 &9.577 \\
AvgJerk\,$\downarrow$  &9.043 &9.128 &\cellcolor{purple2}9.025 &\cellcolor{purple1}8.882 &9.134 \\
IteTime\,$\downarrow$  &202.8 &\cellcolor{purple1}134.5 &194.2 &151.3 &\cellcolor{purple2}150.9 \\
PassTime\,$\downarrow$ &17.551 &15.429 &18.345 &\cellcolor{purple2}12.416 &\cellcolor{purple1}11.278 \\
PassRate\,$\uparrow$   &21.0 &35.0 &42.0 &\cellcolor{purple2}55.0 &\cellcolor{purple1}57.0 \\

\midrule
-&\multicolumn{5}{c}{7 obstacles}\\
AvgAcc\,$\downarrow$   &1.947 &1.748 &2.114 &\cellcolor{purple2}1.591 &\cellcolor{purple1}1.566 \\
MaxAcc\,$\downarrow$   &\cellcolor{purple2}9.688 &9.931 &9.786 &9.779 &\cellcolor{purple1}9.593 \\
AvgJerk\,$\downarrow$  &10.961 &11.935 &9.601 &\cellcolor{purple1}8.974 &\cellcolor{purple2}9.065 \\
IteTime\,$\downarrow$  &234.0 &\cellcolor{purple1}180.9 &260.4 &\cellcolor{purple2}218.9 &222.4 \\
PassTime\,$\downarrow$ &26.710 &22.886 &28.439 &\cellcolor{purple2}18.054 &\cellcolor{purple1}17.441 \\
PassRate\,$\uparrow$   &2.0 &26.0 &30.0 &\cellcolor{purple2}39.0 &\cellcolor{purple1}45.0 \\
\bottomrule
\end{tabular}
\label{tab:performance}
\vspace{-0.1in}
\end{table}

\vspace{-0.12in}
\subsection{Evaluation on Real-World Datasets}\label{sec:exp_dkal}

To evaluate D-KalmanNet's robustness to real-world sensor noises, we conduct experiments on AevaScene~\cite{aevascenes} dataset and compare the following methods:
\textbf{(1) D-KNet}: our proposed D-KalmanNet;
\textbf{(2) KNet}~\cite{revach2022kalmannet}: the KalmanNet, a physics-informed deep learning method; 
\textbf{(3) KF}~\cite{jian2023dynamic}: the Kalman filter, a widely-used conventional tracking method; 
\textbf{(4) D-KF}~\cite{peng2021detection}: the conventional Kalman filter aided by Doppler LiDAR.

To quantify tracking accuracy, we use the normalized mean squared error (NMSE) in decibels (dB) following~\cite{revach2022kalmannet}. Specifically, for each vehicle, and at every feasible time step $t$ , we first evaluate the horizon-averaged NMSE over $\mathcal{H}_t$:
\begin{align}
    \text{NMSE}_t =
        \frac{1}{H} \sum_{h\in\mathcal{H}_t} 
        \frac{(\hat{x}_{h+1|t} - x_{h+1})^2 + (\hat{y}_{h+1|t} - y_{h+1})^2}
        { x_{h+1}^2 + y_{h+1}^2}, \nonumber
\end{align}
where $(\hat{x}_{h+1|t}, \hat{y}_{h+1|t})$ and $(x_{h+1}, y_{h+1})$ denote the predicted and ground-truth obstacle positions, respectively.
We then convert to the logarithmic scale via $10 \log_{10}(\cdot)$
to obtain a per-step dB-NMSE score. These scores are averaged over vehicle trajectory length to yield a single dB-NMSE value per vehicle, and finally aggregated across all vehicles as \textbf{mean $\pm$ std}. 

\begin{table}[t!]
\centering
\caption{Sensitivity Analysis of DT-MPC}
\vspace{-0.05in}
\resizebox{0.48\textwidth}{!}{
\begin{tabular}{ccccccc}
\toprule
\textbf{Parameter}& \multicolumn{2}{c}{3 obstacles} & \multicolumn{2}{c}{5 obstacles} & \multicolumn{2}{c}{7 obstacles} \\
\textbf{Setting}& AvgJerk$\downarrow$ & PassRate$\uparrow$ & AvgJerk$\downarrow$ & PassRate$\uparrow$ & AvgJerk$\downarrow$ & PassRate$\uparrow$\\
\midrule
default &8.058 &88.0 &9.134 &57.0 &9.056 &45.0 \\
\cdashline{1-7}[3pt/3pt]
\addlinespace[3pt]
$\alpha=1.0$ &8.264 &92.0 &9.001 &59.0 &9.077 &44.0 \\
$\alpha=0.5$ &8.179 &89.0 &8.972 &54.0 &8.796 &47.0 \\
$\alpha=0.1$ &7.953 &86.0 &8.799 &48.0 &8.436 &40.0 \\
\cdashline{1-7}[3pt/3pt]
\addlinespace[3pt]
$\beta=1.0$ &8.349 &90.0 &9.206 &61.0 &9.165 &48.0 \\
$\beta=0.5$ &8.370 &79.0 &8.715 &55.0 &9.033 &39.0 \\
$\beta=0.1$ &8.169 &81.0 &8.984 &57.0 &8.912 &43.0 \\
\midrule
\textbf{$\boldsymbol{t}_{\text{DT-MPC}}\,(\text{ms})$} & \multicolumn{2}{c}{0.621} & \multicolumn{2}{c}{0.997} & \multicolumn{2}{c}{1.498} \\
\bottomrule
\end{tabular}
}
\label{tab:sens}
\vspace{-0.02in}
\end{table}

\begin{figure*}[t!]
\centering
\includegraphics[width=0.98\textwidth]{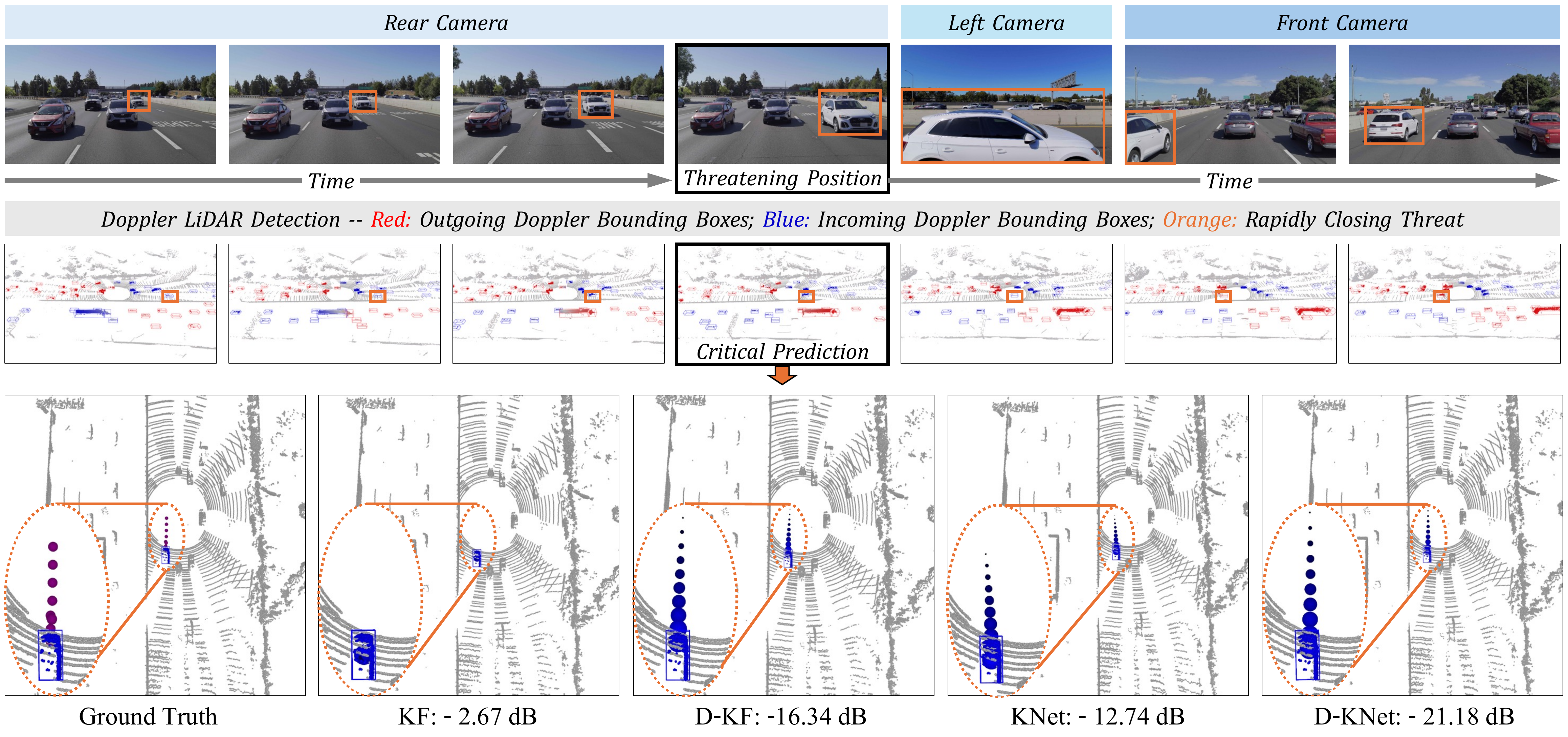}
\vspace{-0.02in}
\caption{Visualization of critical predictions for a rapidly approaching vehicle (at about 25\,m/s, highlighted in orange) in \textit{Highway} scenario of AevaScene~\cite{aevascenes}. The upper half shows Doppler LiDAR's environmental dynamics detection (outgoing\,/\,incoming bounding boxes marked in red\,/\,blue). The lower half provides detailed comparison of trajectory predictions for the threat.}
\label{fig:tracking}
\vspace{-0.1in}
\end{figure*}

\textbf{Qualitative Analysis.}
Fig. \ref{fig:tracking} demonstrates a visualized result of tracking one rapidly approaching vehicle (at about 25\,m/s) on the \textit{Highway} scenario.
Among the evaluated methods, D-KNet demonstrates superior performance by effectively leveraging Doppler cues to enable accurate, real-time tracking of rapid objects, achieving the lowest tracking error of $–21.18$\,dB. In contrast, KF and KNet struggle to handle the approaching vehicle with abrupt accelerations. This comparison highlights the importance of integrating model-based learning and Doppler LiDAR measurements for more robust tracking under real-world settings.

\begin{table}[t]
\centering
\caption{Prediction NMSE (dB) by Mean $\pm$ Std}
\vspace{-0.02in}
\begin{tabular}{ccccc}
\toprule
\multirow{2}{*}{$\mathcal{H}_t$}& \multirow{2}{*}{\textbf{Method}} & \multicolumn{3}{c}{\textbf{Tracking Frequency}}
 \\
\addlinespace[1pt]
&& 10\,Hz & 5\,Hz & 2\,Hz\\

\midrule
\multirow{10}{*}{5} & -& \multicolumn{3}{c}{\textit{Highway} scenario}  \\
\addlinespace[1pt]
&KF \cite{jian2023dynamic} &-22.66 $\pm$ 6.41&	-16.62 $\pm$ 5.24&	-7.18 $\pm$ 3.84 \\
&KNet \cite{revach2022kalmannet} &-23.38 $\pm$ 7.82&	-17.79 $\pm$ 7.77&	-12.56 $\pm$ 7.16 \\
&D-KF \cite{peng2021detection} &-27.15 $\pm$ 6.39&	-21.66 $\pm$ 4.77&	-14.09 $\pm$ 3.60 \\
\cdashline{2-5}[3pt/3pt]
\addlinespace[2pt]
&D-KNet & \cellcolor{purple1}-35.80 $\pm$ 8.62&\cellcolor{purple1}-29.11 $\pm$ 7.17&\cellcolor{purple1}-20.22 $\pm$ 2.45 \\

\cmidrule{2-5}
& -&\multicolumn{3}{c}{\textit{City} scenario}\\
\addlinespace[1pt]
&KF \cite{jian2023dynamic}   &-27.15 $\pm$ 6.35&	-20.48 $\pm$ 4.79&	-9.15 $\pm$ 6.64 \\
&KNet \cite{revach2022kalmannet} &-29.26 $\pm$ 5.47&	-23.66 $\pm$ 5.98&	-17.38 $\pm$ 7.36 \\
&D-KF \cite{peng2021detection} &-32.74 $\pm$ 6.18&	-27.60 $\pm$ 5.11&	-19.20 $\pm$ 5.03 \\
\cdashline{2-5}[3pt/3pt]
\addlinespace[2pt]
&D-KNet &\cellcolor{purple1}-45.00 $\pm$ 7.35&\cellcolor{purple1}	-37.67 $\pm$ 6.11&\cellcolor{purple1}-29.63 $\pm$ 6.76 \\

\midrule
\multirow{10}{*}{10} & - & \multicolumn{3}{c}{\textit{Highway} scenario}  \\
\addlinespace[1pt]
&KF \cite{jian2023dynamic}   &-18.50 $\pm$ 6.53&-11.49 $\pm$ 5.92&	1.88 $\pm$ 6.47\\
&KNet \cite{revach2022kalmannet} & -17.50 $\pm$ 7.49&	-11.83 $\pm$ 7.64&	-6.32 $\pm$ 6.73 \\
&D-KF \cite{peng2021detection} &-23.68 $\pm$ 6.44&	-17.36 $\pm$ 4.65&	-6.61 $\pm$ 4.92 \\
\cdashline{2-5}[3pt/3pt]
\addlinespace[2pt]
&D-KNet &\cellcolor{purple1}-26.28 $\pm$ 8.33&\cellcolor{purple1}-18.88 $\pm$ 6.60&\cellcolor{purple1}-8.56 $\pm$ 4.66\\

\cmidrule{2-5}
&- &\multicolumn{3}{c}{\textit{City} scenario}\\
\addlinespace[1pt]
&KF \cite{jian2023dynamic} &-22.55 $\pm$ 6.46&	-14.14 $\pm$ 5.27&	8.05 $\pm$ 0.30 \\ 
&KNet \cite{revach2022kalmannet} &-23.19 $\pm$ 5.54&	-17.34 $\pm$ 6.12&	-9.81 $\pm$ 8.92\\
&D-KF \cite{peng2021detection} &-29.67 $\pm$ 6.90&	-22.97 $\pm$ 5.59&	-10.59 $\pm$ 7.26\\
\cdashline{2-5}[3pt/3pt]
\addlinespace[2pt]
&D-KNet &\cellcolor{purple1}-34.30 $\pm$ 7.65&\cellcolor{purple1}-25.96 $\pm$ 6.82&\cellcolor{purple1}-16.41 $\pm$ 9.47\\

\bottomrule
\end{tabular}
\label{tab:tracking}
\vspace{-0.1in}
\end{table}

\textbf{Quantitative Analysis.}
Table \ref{tab:tracking} reports the results across tracking frequencies (2\,-\,10 Hz), prediction horizons ($\mathcal{H}_t=5,10$), and driving scenarios (\textit{Highway}, \textit{City}). Across all settings, D-KNet consistently achieves the lowest NMSE and significantly outperforms all baselines. Notably, at $\mathcal{H}_t=5$, D-KNet attains $-45.00 \pm 7.35$ dB in the \textit{City} scenario at 10 Hz and surpasses the second best D-KF~\cite{peng2021detection} by $12.26$ dB. 
Even at $\mathcal{H}_t=10$, D-KNet maintains robustness to long-horizon uncertainty, whereas other methods suffer considerable degradation. Tighter standard deviations of D-KNet further confirm its stability across diverse obstacle dynamics. These results demonstrate the superior reliability of D-KNet, especially under real-world constraints such as varying sensor noise, limited processing frequency, and scenario complexity. 

\vspace{-0.15in}
\subsection{More Quantitative Results}

We also deploy D-KNet on an NVIDIA Jetson Orin NX 16\,GB to evaluate its real-time tracking efficiency on a resource-constrained platform. 
It is found that D-KNet consumes only about $107~\text{MB}$ of GPU memory. 
As Table \ref{tab:orin} presents, it achieves inference rates of over $100$\,Hz when tracking a single fast-moving obstacle. Notably, even with $10$ simultaneously tracked obstacles, D-KNet sustains about $15$\,Hz with low CPU usage of $42.0\,\%$. These confirm D-KNet’s suitability for lightweight deployment on robot platforms.

Table~\ref{tab:horizon} shows per-step NMSE (dB) by \textbf{mean} $\pm$ \textbf{std} over a 10-step prediction horizon on the AevaScenes~\cite{aevascenes} dataset. 
D-KNet consistently outperforms all baselines at every step, with margin widening especially in short steps (e.g., $-48.81$ dB of DKNet and $-16.49$ dB of KF for step $1$ in \textit{City}). 
The KF and D-KF methods exhibit nearly flat error growth, indicating trivial estimates under transition model mismatch. However, D-KNet and KNet show physically reasonable error accumulation, demonstrating the effectiveness of model-based learning against dynamics uncertainty. 
Notably, D-KNet maintains tighter standard deviations and significantly lower mean error, verifying that the Doppler LiDAR integration enhances both accuracy and robustness for long-horizon predictions.

\begin{table}[t]
\centering
\caption{Hardware Efficiency of D-KNet}
\vspace{-0.02in}
\resizebox{0.49\textwidth}{!}{
\begin{tabular}{cccccccc}
\toprule
 \textbf{Obstacle Number}& 1 & 2 & 3 & 4 & 5 & 8 & 10 \\
\midrule
Execution Time (s)      & 0.008 & 0.016 & 0.025& 0.032 & 0.037 & 0.050 & 0.070\\
Frequency (Hz)          & 126.57 & 62.08 & 39.93& 31.53 & 27.10 & 19.87 & 14.36 \\
CPU Usage (\%)          & 5.0    & 8.0     & 27.0    & 38.0    & 40.0   & 41.0   & 42.0   \\
\bottomrule
\end{tabular}
}
\label{tab:orin}
\vspace{-0.1in}
\end{table}

\begin{figure}[t]
\centering
    \begin{subfigure}[t]{0.15\textwidth}
      \includegraphics[width=\textwidth]{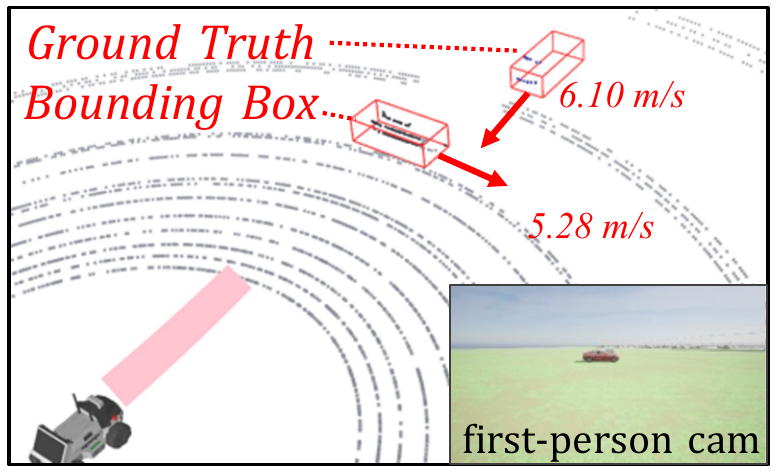}
      \caption{Precrash.}
    \end{subfigure}
    \hspace{-1.5mm}
    \begin{subfigure}[t]{0.15\textwidth}
        \includegraphics[width=1.00\textwidth]{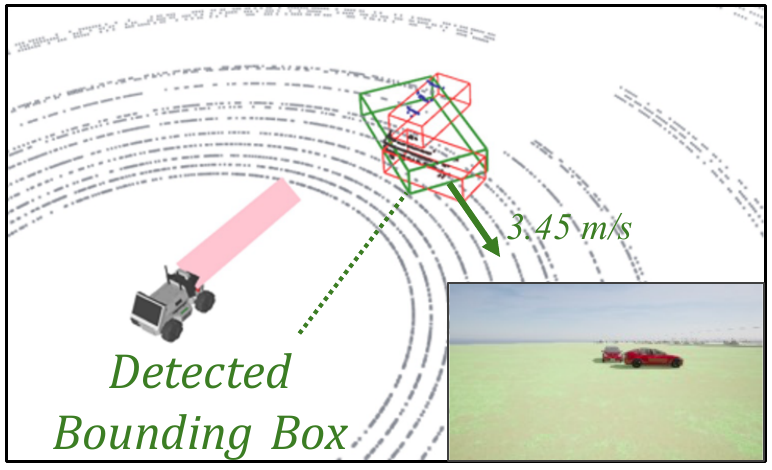}
        \caption{Hallucination.}
    \end{subfigure}
    \hspace{-1.5mm}
    \begin{subfigure}[t]{0.15\textwidth}
        \includegraphics[width=1.00\textwidth]{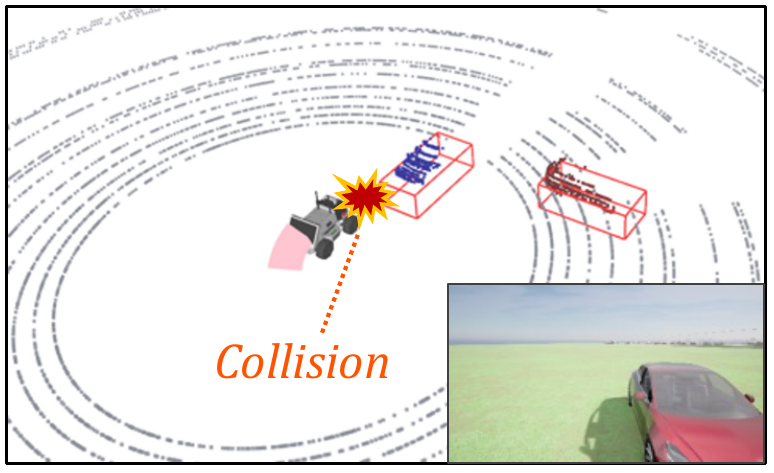}
        \caption{Crash.}
    \end{subfigure}
    \vspace{-0.02in}
    \caption{
    Failure mode analysis.
    }
    \label{fig:f_mode}
    \vspace{-0.2in}
\end{figure}

\begin{table*}[t!]
\label{horzion}
\centering
\caption{Per-horizon-step Prediction NMSE (dB) by Mean $\pm$ Std}
\vspace{-0.02in}
\resizebox{1.00\textwidth}{!}{
\begin{tabular}{ccccccccccc}
\toprule
\multirow{2}{*}{\textbf{Method}} & \multicolumn{10}{c}{\textbf{Horizon Step}}
 \\
\addlinespace[1pt]
& 1 & 2 & 3&4&5&6&7&8&9&10 \\

\midrule
 -& \multicolumn{10}{c}{\textit{Highway} scenario}  \\
\addlinespace[1pt]
KF \cite{jian2023dynamic}   &-15.53 $\pm$ 8.80	&	-15.15 $\pm$ 9.63	&	-14.72 $\pm$ 8.45	&	-14.24 $\pm$ 8.26	&	-13.73 $\pm$ 8.09	&	-13.20 $\pm$ 7.90	&	-12.66 $\pm$ 7.72	&	-12.08 $\pm$ 7.53	&	-11.52 $\pm$ 7.33	&	-10.96 $\pm$ 7.15\\
KNet \cite{revach2022kalmannet} &-30.82 $\pm$ 3.80	&	-24.80 $\pm$ 4.05 	&	-21.44 $\pm$ 4.28 	&	-18.93 $\pm$ 4.50 	&	-16.93 $\pm$ 4.70 	&	-15.24 $\pm$ 4.87 	&	-13.78 $\pm$ 5.03 	&	-12.42 $\pm$ 5.19 	&	-11.23 $\pm$ 5.34 	&	-10.16 $\pm$ 5.50\\
D-KF \cite{peng2021detection} & -16.10 $\pm$ 10.37	&	-16.01 $\pm$ 9.51	&	-15.88 $\pm$ 9.19	&	-15.74 $\pm$ 8.91	&	-15.56 $\pm$ 8.64	&	-15.36 $\pm$ 8.47	&	-15.13 $\pm$ 8.33	&	-14.88 $\pm$ 8.23	&	-14.58 $\pm$ 8.14	&	\cellcolor{purple1}-14.27 $\pm$ 8.10\\
\cdashline{2-11}[3pt/3pt]
\addlinespace[2pt]
D-KNet &\cellcolor{purple1}-41.39 $\pm$ 3.23 	&	\cellcolor{purple1}-34.79 $\pm$ 3.34 	&	\cellcolor{purple1}-30.17 $\pm$ 3.47 	&	\cellcolor{purple1}-26.63 $\pm$ 3.58 	&	\cellcolor{purple1}-23.74 $\pm$ 3.73 	&	\cellcolor{purple1}-21.27 $\pm$ 3.90 	&	\cellcolor{purple1}-19.14 $\pm$ 4.10	&	\cellcolor{purple1}-17.23 $\pm$ 4.29	&	\cellcolor{purple1}-15.45 $\pm$ 4.52 	&	-13.89 $\pm$ 4.74  \\

\cmidrule{1-11}
-&\multicolumn{10}{c}{\textit{City} scenario}\\
\addlinespace[1pt]
KF \cite{jian2023dynamic}   &-16.49 $\pm$ 7.12	&	-16.34 $\pm$ 6.83	&	-16.32 $\pm$ 6.69	&	-16.15 $\pm$ 6.54	&	-15.92 $\pm$ 6.42	&	-15.65 $\pm$ 6.30	&	-15.34 $\pm$ 6.17	&	-14.98 $\pm$ 6.04	&	-14.59 $\pm$ 5.91	&	-14.16 $\pm$ 5.77\\
KNet \cite{revach2022kalmannet} &-38.22 $\pm$ 3.05 	&	-32.42 $\pm$ 3.10 	&	-28.90 $\pm$ 3.16 	&	-26.26 $\pm$ 3.42 	&	-24.13 $\pm$ 3.57 	&	-22.3 $\pm$ 3.72 	&	-20.68 $\pm$ 3.88 	&	-19.22 $\pm$ 3.96 	&	-17.87 $\pm$ 3.95	&	-16.59 $\pm$ 4.06  \\
D-KF \cite{peng2021detection} &-16.29 $\pm$ 8.89	&	-16.26 $\pm$ 8.19	&	-16.23 $\pm$ 7.78	&	-16.19 $\pm$ 7.47	&	-16.17 $\pm$ 7.26	&	-16.13 $\pm$ 7.07	&	-16.13 $\pm$ 6.88	&	-16.13 $\pm$ 6.68	&	-16.12 $\pm$ 6.51	&	-16.11 $\pm$ 6.34\\
\cdashline{2-11}[3pt/3pt]
\addlinespace[2pt]
D-KNet &\cellcolor{purple1}-48.81 $\pm$ 3.01 	&	\cellcolor{purple1}-41.47 $\pm$ 3.09	&	\cellcolor{purple1}-36.48 $\pm$ 3.21	&	\cellcolor{purple1}-32.63 $\pm$ 3.25 	&	\cellcolor{purple1}-29.44 $\pm$ 3.17 	&	\cellcolor{purple1}-27.76 $\pm$ 3.31 	&	\cellcolor{purple1}-24.45 $\pm$ 3.57 	&	\cellcolor{purple1}-22.41 $\pm$ 3.62	&	\cellcolor{purple1}-20.59 $\pm$ 3.65 	&	\cellcolor{purple1}-18.95 $\pm$ 3.76  \\

\bottomrule
\end{tabular}
}
\label{tab:horizon}
\vspace{-0.15in}
\end{table*}

\vspace{-0.15in}
\subsection{Limitation and Future Work}

In practice, bounding box detection noises (e.g., jitters, false positives) may affect obstacle linear velocity estimation and then downstream planning. 
Fig.~\ref{fig:f_mode} demonstrates a failure mode case where a severe box under-segmentation occurs when two rapid obstacles (moving at $6.10\,$m/s and $5.28\,$m/s) approach, as depicted in (b).
This contaminates linear velocity estimation and obstacle tracking with a hallucinated object moving at $3.45\,$m/s and propagates to downstream planning, resulting in a collision failure eventually, shown in (c).
To address this, our future work aims to develop bounding-box-free methods for Doppler LiDAR obstacle tracking.
In terms of controller parameter tuning, DT-MPC explores Doppler-guided tuning with a heuristic strategy, which is in general suboptimal for minimizing $\mathcal{L}$, though proved to be efficient and effective.
To address this, our future work aims to integrate gradient-informed techniques like differentiable optimization or policy search~\cite{cheng2024difftune,song2022policy} for optimal Doppler-guided tuning, while maintaining model efficiency.

\vspace{-0.05in}
\section{Conclusion}

This paper has introduced DPNet, a Doppler LiDAR model-based learning method for collision avoidence among rapid-motion obstacles.
DPNet achieves both high frequency and accuracy in tracking and planning. 
Compared to benchmark schemes, it reduces the navigation time by $6\,\%$--$30\,\%$, increases the success rate by up to $16\,\%$, and reduces the tracking error by over $10$\,dB.
It was demonstrated that DPNet adapts to platforms with limited computational resources.
Ablation studies confirmed the indispensability of incorporating Doppler information into controller tuning and motion planning.

\vspace{-0.05in}

\bibliographystyle{IEEEtran}
\bibliography{main}
\end{document}